\documentclass[letterpaper]{article} 
\usepackage{aaai25}  
\usepackage{times}  
\usepackage{helvet}  
\usepackage{courier}  
\usepackage[hyphens]{url}  
\usepackage{graphicx} 
\usepackage{natbib}  
\usepackage{caption} 
\usepackage{algorithm}
\usepackage{algorithmic}
\usepackage{amsmath}

\usepackage{graphicx}
\usepackage[table,xcdraw]{xcolor}

\usepackage{enumitem}
\usepackage{booktabs}
\usepackage{multirow}
\usepackage{newfloat}
\usepackage{listings}
\DeclareCaptionStyle{ruled}{labelfont=normalfont,labelsep=colon,strut=off} 
\floatstyle{ruled}
\newfloat{listing}{tb}{lst}{}
\floatname{listing}{Listing}

\title{A Dataset for Analysing News Framing in Chinese Media}

\author {
    Owen Cook\equalcontrib,
    Yida Mu\equalcontrib,
    Xinye Yang\equalcontrib,
    Xingyi Song
    and
    Kalina Bontcheva
}
\affiliations {
    School of Computer Science, University of Sheffield\\
    \{oscook1, x.song, k.bontcheva\}@sheffield.ac.uk
}

\begin{document}

\maketitle

\begin{abstract}
Framing is an essential device in news reporting, allowing the writer to influence public perceptions of current affairs. While there are existing automatic news framing detection datasets in various languages, none of them focus on news framing in the Chinese language which has complex character meanings and unique linguistic features. This study introduces the first Chinese News Framing dataset, to be used as either a stand-alone dataset or a supplementary resource to the SemEval-2023 task 3 dataset. We detail its creation and we run baseline experiments to highlight the need for such a dataset and create benchmarks for future research, providing results obtained through fine-tuning XLM-RoBERTa-Base and using GPT-4o in the zero-shot setting. We find that GPT-4o performs significantly worse than fine-tuned XLM-RoBERTa across all languages. For the Chinese language, we obtain an F1-micro (the performance metric for SemEval task 3, subtask 2) score of 0.719 using only samples from our Chinese News Framing dataset and a score of 0.753 when we augment the SemEval dataset with Chinese news framing samples. With positive news frame detection results, this dataset is a valuable resource for detecting news frames in the Chinese language and is a valuable supplement to the SemEval-2023 task 3 dataset.

\end{abstract}

\section{Introduction}
Framing is a fundamental process for understanding the world. It is done when people both communicate and receive information \citep{entman1993framing} and it shapes a person’s conceptualisation of the things around them. While scholars are not in agreement about an exact definition \citep{hertog2001multiperspectival, van2023analyzing}, we refer to that of \citet{gamson1992media}, that a frame is ``a central organising principle that holds together and gives coherence and meaning to a diverse array of symbols''. A symbol may be anything from a small facial expression to a group of words that are commonly used to evoke particular emotions. The interpretation of symbols by a receiver is influenced by their existing knowledge and internal grouping of concepts, impacting their perception of information. The communicator also frames the symbols they are projecting, allowing them to influence the framing of the receiver.

In the news and media, framing is omnipresent; social media content creators, journalists, and politicians particularly influence how the public view and feel about current affairs. \citet{d2018doing} refers to this as news framing (also known as media framing): ``how journalists, their sources, and audiences work within conditions that shape the messages they construct as well as the ways they understand and interpret these messages''.

Due to the vastness of information available online, there is a desire for automatic news frame detection. Automatic news frame detection is the use of computational methods to detect news frames. It has a variety of applications, such as understanding media bias \citep{morstatter2018identifying}, providing balanced framing of articles at the information retrieval stage \citep{reiter2024framefinder}, automating large-scale content analysis \citep{kwak2020systematic, alonso2023framing}, and detecting misinformation \citep{wang2024framedtruth}. Datasets available for the development of language models capable of automatically detecting news frames span a number of languages, yet news framing in the Chinese language is particularly under-explored.

This work introduces the Chinese News Framing dataset, facilitating research investigating the training of models capable of detecting news frames in Chinese. The dataset has been designed such that it also acts as a complementary resource for the SemEval-2023 task 3 dataset \citep{piskorski-etal-2023-semeval}. We provide baseline experiments highlighting the performance of the multilingual language model XLM-RoBERTa \citep{conneau2019unsupervised} on Chinese news framing samples with fine-tuning on the SemEval training set, our Chinese News Framing training set, and our augmented SemEval training (a concatenation of both training sets). We also provide results obtained by GPT-4o in the zero-shot setting for all test samples in each language. We make our dataset and code available on Zenodo \footnote{Dataset: \url{https://doi.org/10.5281/zenodo.14659362}}, and GitHub \footnote{Code: \url{https://github.com/GateNLP/chinese-news-framing}} with the CC-BY-NC-SA License.

\section{Related Work}

\subsection{News Frames}
News framing can emphasise certain aspects of an issue, shaping public perception and interpretation \citep{de2005news,lecheler2019news} through the selective emphasis and organisation of specific elements within news stories.

By highlighting certain aspects of events while downplaying others, frames can profoundly influence how people think about issues, attribute responsibility, and evaluate potential solutions \citep{Lecheler2012News}. The impact of news frames manifests itself in multiple aspects. Frame analysis reveals how media organisations in different countries and languages present the same events through different cultural and ideological perspectives. Understanding these framing patterns also enables researchers to track how public discourse evolves over time on global issues such as climate change, public health, and international conflicts \citep{card-etal-2015-media}.

\subsection{News Framing Detection}
In an era of increasing digital news consumption, the ability to detect and analyse frames systematically has become essential for understanding media influence and bias across diverse platforms and contexts \citep{hamborg2019automated}. To detect news frames, a number of approaches have been employed. These include topic modelling \citep{dimaggio2013exploiting}, hierarchical topic modelling \citep{nguyen2015tea}, cluster and sentiment analysis \citep{burscher2016frames}, and more recently neural network models, with transformer-based models now highly performative and widely used \citep{liu2019detecting, akyurek2020multi, piskorski-etal-2023-semeval}. One of the most commonly used models in the SemEval-2023 task 3 was the multilingual BERT-based model XLM-RoBERTa \citep{wu2023sheffieldveraai, jiang-2023-team, liao-etal-2023-marseclipse}.

To achieve the automatic detection of news framing, \citet{card2015media} developed the first large-scale dataset ``Media Frames Corpus'' and the corresponding annotation framework. The Media Frames Corpus consists of 20k English articles annotated into one or more frames from 15 categories introduced by \citet{boydstun2014tracking}; the topics in this dataset include immigration, smoking, and same-sex marriage. \citet{liu2019detecting} introduced the Gun Violence Frame Corpus (GVFC) focussing solely on gun violence and the English language, using a different set of frames. Utilising the same frame definitions as \citet{card-etal-2015-media}, \citet{piskorski-etal-2023-semeval} introduced the first multilingual news framing dataset (SemEval-2023 task 3); the publicly available training and development sets contain samples in English, French, German, Italian, Polish, and Russian.

The lack of Chinese content in these existing datasets have clear limitations for Chinese news framing analysis. The unique linguistic features of the Chinese language, such as the lack of explicit word boundaries and complex character-meanings and relationships, pose special challenges. To the best of our knowledge, a Chinese dataset for automatic news framing detection is yet to be developed. To address these gaps, we introduce the first Chinese News Framing dataset. This resource allows for the analysis of framing patterns across different types of Chinese media discourse across a variety of news sources.

\section{Task Description}
For a given news article, determine one or more frames applied in the text from a set of 14 generic framing dimensions \citep{card2015media,piskorski-etal-2023-semeval}: (1) Economic,
(2) Capacity and Resources, (3) Morality, (4) Fairness and Equality, (5) Legality, Constitutionality and Jurisprudence, (6) Policy Prescription and Evaluation, (7) Crime and Punishment, (8) Security and Defence, (9) Health and Safety, (10) Quality of Life, (11) Cultural Identity, (12) Public Opinion, (13) Political, and (14) External Regulation and Reputation. We frame our task as a \textbf{multi-class multi-label classification} problem at the news article level. Table \ref{tab:label_and_definitions} lists all the framework dimensions and their corresponding definitions.

\begin{table*}[!t]
\resizebox{\textwidth}{!}{%
\begin{tabular}{ll}
\hline
\rowcolor[HTML]{EFEFEF} 
\multicolumn{1}{c}{\cellcolor[HTML]{EFEFEF}\textbf{Category}} &
  \multicolumn{1}{c}{\cellcolor[HTML]{EFEFEF}\textbf{Definition}} \\ \hline
\textbf{\begin{tabular}[c]{@{}l@{}}1: Economic\end{tabular}} &
  \begin{tabular}[c]{@{}l@{}}This type identifies parts of the articles referring to costs, benefits, or other financial\\ implications.\end{tabular} \\ \cline{2-2} 
\textbf{\begin{tabular}[c]{@{}l@{}}2: Capacity and Resources\end{tabular}} &
  \begin{tabular}[c]{@{}l@{}}This type identifies parts of the articles referring to the availability of physical, human,\\ or financial resources, and the capacity of current systems.\end{tabular} \\ \cline{2-2} 
\textbf{\begin{tabular}[c]{@{}l@{}}3: Morality\end{tabular}} &
  \begin{tabular}[c]{@{}l@{}}This type identifies parts of the articles referring to religious or ethical implications.\end{tabular} \\ \cline{2-2} 
\textbf{\begin{tabular}[c]{@{}l@{}}4: Fairness and Equality\end{tabular}} &
  \begin{tabular}[c]{@{}l@{}}This type identifies parts of the articles referring to the balance or distribution of rights,\\ responsibilities, and resources.\end{tabular} \\ \cline{2-2} 
\textbf{\begin{tabular}[c]{@{}l@{}}5: Legality, Constitutionality and\\ Jurisprudence\end{tabular}} &
  \begin{tabular}[c]{@{}l@{}}This type identifies parts of the articles referring to rights, freedoms, and authority of\\ individuals, corporations, and government.\end{tabular} \\ \cline{2-2} 
\textbf{\begin{tabular}[c]{@{}l@{}}6: Policy Prescription and Evaluation\end{tabular}} &
  \begin{tabular}[c]{@{}l@{}}This type identifies parts of the articles referring to discussion of specific policies aimed\\ at addressing problems.\end{tabular} \\ \cline{2-2} 
\textbf{\begin{tabular}[c]{@{}l@{}}7: Crime and Punishment\end{tabular}} &
  \begin{tabular}[c]{@{}l@{}}This type identifies parts of the articles referring to the effectiveness and the implica-\\ tions of laws and their enforcement.\end{tabular} \\ \cline{2-2} 
\textbf{\begin{tabular}[c]{@{}l@{}}8: Security and Defence\end{tabular}} &
  \begin{tabular}[c]{@{}l@{}}This type identifies parts of the articles referring to threats to welfare of the individual,\\ community, or nation.\end{tabular} \\ \cline{2-2} 
\textbf{\begin{tabular}[c]{@{}l@{}}9: Health and Safety\end{tabular}} &
  \begin{tabular}[c]{@{}l@{}}This type identifies parts of the articles referring to health care, sanitation, and public\\ safety\end{tabular} \\ \cline{2-2} 
\textbf{\begin{tabular}[c]{@{}l@{}}10: Quality of Life\end{tabular}} &
  \begin{tabular}[c]{@{}l@{}}This type identifies parts of the articles referring to threats and opportunities for the\\ individual's wealth, happiness, and well-being.\end{tabular} \\ \cline{2-2} 
\textbf{\begin{tabular}[c]{@{}l@{}}11: Cultural Identity\end{tabular}} &
  \begin{tabular}[c]{@{}l@{}}This type identifies parts of the articles referring to traditions, customs, or values of a\\ social group in relation to a policy issue.\end{tabular} \\ \cline{2-2} 
\textbf{\begin{tabular}[c]{@{}l@{}}12: Public Opinion\end{tabular}} &
  \begin{tabular}[c]{@{}l@{}}This type identifies parts of the articles referring to attitudes and opinions of the general\\ public, including polling and demographics.\end{tabular} \\ \cline{2-2} 
\textbf{\begin{tabular}[c]{@{}l@{}}13: Political\end{tabular}} &
  \begin{tabular}[c]{@{}l@{}}This type identifies parts of the articles referring to considerations related to politics\\ and politicians, including lobbying, elections, and attempts to sway voters.\end{tabular} \\ \cline{2-2} 
\textbf{\begin{tabular}[c]{@{}l@{}}14: External Regulation and Reputation\end{tabular}} &
  \begin{tabular}[c]{@{}l@{}}This type identifies parts of the articles referring to international reputation or foreign\\ policy.\end{tabular} \\ \hline
\end{tabular}
}
\caption{Chinese news framing categories and definitions as used in \citet{piskorski-etal-2023-semeval}.}
\label{tab:label_and_definitions}
\end{table*}

\section{Data}
To annotate frames in Chinese articles, we use the pipeline adopted by \citet{piskorski-etal-2023-semeval} which has been successfully applied to various languages. Specifically, our dataset development framework is divided into three distinct steps:

\begin{itemize} 
\item \textit{Data Collection.} To support our document-level annotation task, we begin by gathering a dataset of Chinese news articles $T$.
\item \textit{Data Sampling.} From $T$, we select a representative subset of articles $D$ for annotation.
\item \textit{Data Annotation.} Lastly, we provide a detailed description of the annotation process applied to $D$. 
\end{itemize}

\begin{table}[!t]
{
\begin{tabular}{lll}
\hline
\rowcolor[HTML]{EFEFEF} 
\textbf{New Sources}  & \textbf{Country} & \textbf{\# of articles} \\ \hline
BBC Chinese                        & UK      & 3k   \\
Voice of America             & USA     & 36k  \\
Financial Times              & UK      & 3k  \\
Reuters                            & UK      & 23k  \\
Deutsche Welle                & Germany & 24k  \\
China Digital Times           &  USA & 14k  \\
The New York Times Chinese    &  USA  & 6k   \\
Radio France Internationale   & France  & 3k   \\
Radio Free Asia              &  USA     & 11k  \\ 
New Tang Dynasty                   &  USA      & 79k  \\
The Epoch Times                    &  USA      & 100k \\
{  China Daily}                    &  China     & 3k   \\
{  The Paper}                      &  China    & 2k   \\
{  Xinhua News Agency}             &  China    & 1k   \\ \hline
\end{tabular}
}
\caption{Chinese news outlets we collected data from. The ``Country'' column indicates the media outlet’s base location.}
\label{tab:news_source}
\end{table}

\subsection{Data Collection}
\noindent \textbf{News sources} \\
Following \citet{piskorski-etal-2023-semeval}, we collected Chinese news articles from 13 different websites spanning 5 countries. We also selected these news outlets to ensure a balance of political biases, as determined by the Media Bias Fact Check (MBFC) platform.\footnote{\url{https://mediabiasfactcheck.com/}} Table \ref{tab:news_source} presents the specifications of each news outlet according to MBFC. \\

\noindent \textbf{Time and Topics} \\
We collect Chinese news articles published between 2020 and the end of 2024. Our collection consists of various globally discussed events, including the COVID-19 vaccine, the Israeli–Palestinian conflict, the Russo-Ukrainian war, and the US election. In total, we obtain approximately 300k news articles across 13 different news sites.

\begin{figure}[!ht]
    \centering
    \includegraphics[width=0.99\columnwidth]{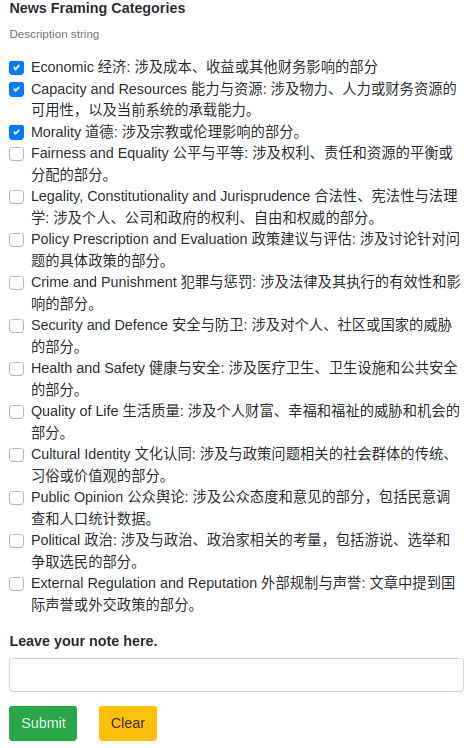}
\caption{GATE Teamware user interface. Annotators can leave comments (see the bottom of the figure), which are used by the expert annotator during the data adjudication process (See Section \textit{Data Adjudication}).}
    \label{fig:gate}
\vspace{-0.5cm}
\end{figure}

\subsection{Data Sampling}

To ensure class balance in the final annotated dataset, we subsample the collected news articles based on their topics. Specifically,given the news corpus \( T = \{t_1, t_2, \dots, t_n\} \), we first employ BERTopic \citep{grootendorst2022bertopic} to assign a primary topic to each article ($t_i$). Next, we compute the cosine similarity between the BERTopic-generated topic representation and each framing category, then pre-assign the article to the provisional framing category ($f^*_j$) based on the similarity between BERTTopic label and defined framing category. The BERTopic-generated topic representation and each framing category are encoded by using the Sentence Transformer\footnote{\url{https://huggingface.co/sentence-transformers/paraphrase-multilingual-MiniLM-L12-v2}} \citep{reimers-2020-multilingual-sentence-bert}.

\begin{align}
    \mathbf{z}_i = \text{SentenceTransformer}(\text{BERTopic}(t_i))\\ 
    \mathbf{f}_j = \text{SentenceTransformer}(f_j),
\end{align}

\noindent Where $z_i$ is embedding of BERT topic of document $t_i$, and $f_j$ is embedding of framing category $j$ \( F = \{f_1, f_2, \dots, f_k\} \).
\(\forall t_i \in T, f_j \in F.\) \\

\noindent \textbf{Framing Assignment} \\
Assign the most likely framing category \( f_j^* \) to each article \( t_i \) by maximizing the cosine similarity between the topic embedding \( \mathbf{z}_i \) and the framing category vectors \( \mathbf{f}_j \):

\begin{equation}
    f_j^* = \arg\max_{f_j} \frac{\mathbf{z}_i \cdot \mathbf{f}_j}{\|\mathbf{z}_i\| \|\mathbf{f}_j\|}.
\end{equation} \\

\noindent \textbf{Stratified Sampling} \\
Based on the pre-assigned topics,
we use a stratified method to sample a subset of 400 news articles ($D$) which is used for the training session and final annotation. 
Note that we consider a size of 400, which is similar to the number of articles per language in \citet{piskorski-etal-2023-semeval}.

\begin{table}[!t]
\small
\begin{tabular}{lllll}
\hline
\rowcolor[HTML]{EFEFEF} 
\multicolumn{1}{c}{\cellcolor[HTML]{EFEFEF}\textbf{}} & \textbf{Train set} & \textbf{Dev set} & \textbf{Test set} & \textbf{All} \\ \hline
\multicolumn{5}{c}{\textbf{Statistics}}                                       \\ \hline
\cellcolor[HTML]{EFEFEF}\textbf{\# of Samples}      & 233  & 50  & 70  & 353  \\
\cellcolor[HTML]{EFEFEF}\textbf{\#avg. Frames}      & 3.17 & 3.0 & 3.1 & 3.13 \\
\cellcolor[HTML]{EFEFEF}\textbf{\#Avg. Tokens}      & 484  & 487 & 469 & 481  \\ \hline
\cellcolor[HTML]{EFEFEF}\textbf{BBC Chinese}        & 2    & 1   & 3   & 6    \\
\cellcolor[HTML]{EFEFEF}\textbf{VOA}                & 35   & 5   & 8   & 48   \\
\cellcolor[HTML]{EFEFEF}\textbf{Reuters}            & 21   & 8   & 7   & 36   \\
\cellcolor[HTML]{EFEFEF}\textbf{NYT Chinese}        & 3    & 1   & 1   & 5    \\
\cellcolor[HTML]{EFEFEF}\textbf{DW}                 & 9    & 3   & 1   & 13   \\
\cellcolor[HTML]{EFEFEF}\textbf{FT}                 & 3    & 2   & 1   & 6    \\
\cellcolor[HTML]{EFEFEF}\textbf{RFA}                & 3    & 1   & 1   & 5    \\
\cellcolor[HTML]{EFEFEF}\textbf{RFI}                & 33   & 6   & 10  & 49   \\
\cellcolor[HTML]{EFEFEF}\textbf{CDT}                & 2    & 1   & 1   & 4    \\
\cellcolor[HTML]{EFEFEF}\textbf{Epoch Times}        & 34   & 9   & 11  & 54   \\
\cellcolor[HTML]{EFEFEF}\textbf{New Tang Dynasty}   & 50   & 3   & 16  & 69   \\
\cellcolor[HTML]{EFEFEF}\textbf{The Paper}          & 11   & 3   & 4   & 18   \\
\cellcolor[HTML]{EFEFEF}\textbf{Xinhua News}        & 11   & 5   & 4   & 20   \\
\cellcolor[HTML]{EFEFEF}\textbf{China Daily}        & 14   & 2   & 4   & 20   \\ \hline
\multicolumn{5}{c}{\textbf{Time}}                                             \\ \hline
\cellcolor[HTML]{EFEFEF}\textbf{2024}               & 39   & 6   & 16  & 61   \\
\cellcolor[HTML]{EFEFEF}\textbf{2023}               & 54   & 11  & 12  & 77   \\
\cellcolor[HTML]{EFEFEF}\textbf{2022}               & 42   & 13  & 18  & 73   \\
\cellcolor[HTML]{EFEFEF}\textbf{2021}               & 47   & 14  & 8   & 69   \\
\cellcolor[HTML]{EFEFEF}\textbf{2020}               & 51   & 6   & 16  & 73   \\ \hline
\multicolumn{5}{c}{\textbf{\# of Samples per Category}}                       \\ \hline
\cellcolor[HTML]{EFEFEF}\textbf{1: Economic}           & 71   & 19  & 21  & 111  \\
\cellcolor[HTML]{EFEFEF}\textbf{2: Capacity}           & 70   & 15  & 17  & 102  \\
\cellcolor[HTML]{EFEFEF}\textbf{3: Morality}           & 22   & 6   & 7   & 35   \\
\cellcolor[HTML]{EFEFEF}\textbf{4: Fairness}           & 27   & 8   & 7   & 42   \\
\cellcolor[HTML]{EFEFEF}\textbf{5: Legality}           & 58   & 9   & 8   & 75   \\
\cellcolor[HTML]{EFEFEF}\textbf{6: Policy}             & 114  & 26  & 28  & 168  \\
\cellcolor[HTML]{EFEFEF}\textbf{7: Crime}              & 57   & 7   & 22  & 86   \\
\cellcolor[HTML]{EFEFEF}\textbf{8: Security}           & 75   & 15  & 23  & 113  \\
\cellcolor[HTML]{EFEFEF}\textbf{9: Health}             & 42   & 5   & 15  & 62   \\
\cellcolor[HTML]{EFEFEF}\textbf{10: Life}               & 41   & 10  & 10  & 61   \\
\cellcolor[HTML]{EFEFEF}\textbf{11: Cultural}           & 25   & 4   & 10  & 39   \\
\cellcolor[HTML]{EFEFEF}\textbf{12: Public Opinion}     & 33   & 4   & 11  & 48   \\
\cellcolor[HTML]{EFEFEF}\textbf{13: Political}          & 28   & 4   & 13  & 45   \\
\cellcolor[HTML]{EFEFEF}\textbf{14: External}           & 77   & 18  & 25  & 120  \\ \hline
\end{tabular}
\caption{Statistics of three subsets.}
\label{tab:dataset_splits}
\vspace{-0.5cm}
\end{table}

\subsection{Data Annotation}

\textbf{Annotation Tool} \\
We annotate Chinese news articles using an open-source data annotation tool, GATE Teamware\footnote{\url{https://github.com/GateNLP/gate-teamware}} \citep{wilby2023gate}, which has been employed in similar computational social science annotation tasks \citep{mu2023vaxxhesitancy,wu2023don,cook2024efficient}. \\

\noindent \textbf{Annotator Training} \\
For annotation, we hired six native-Chinese-speaking undergraduate and postgraduate students from The University of Sheffield at a rate of £17 per hour. We first provided a two-hour training session for all participants, during which we: (i) introduced the task description (See Section Task Description), label definitions (See Table \ref{tab:label_and_definitions}), and guidelines for using GATE Teamware (See Figure \ref{fig:gate}), and (ii) asked all participants to annotate 20 samples. Following this, the senior annotator released the gold-standard labels and explained their decision on each article. The gold-standard labels for the annotator training set were created and validated by three expert annotators, all of whom are native Chinese speakers and experienced NLP researchers with expertise in media analysis.

All annotators were provided with an information sheet containing details about the task and signed consent forms in accordance with ethical guidelines. \\

\noindent \textbf{Task Allocation} \\
Following \citet{piskorski-etal-2023-semeval}, in the initial annotation stage, each news article was annotated by two different annotators. To distribute the samples between six annotators such that each sample has two unique annotators, we employed a modified version of the EffiARA annotation framework \citep{cook2024efficient}. Through using the EffiARA framework, we were able to evenly distribute samples among annotators while maintaining the ability to assess inter- and intra-annotator agreement as well as calculate a combined ``annotator reliability factor'' for each annotator. 

\begin{figure}[!ht]
    \centering
    \includegraphics[width=0.71\columnwidth]{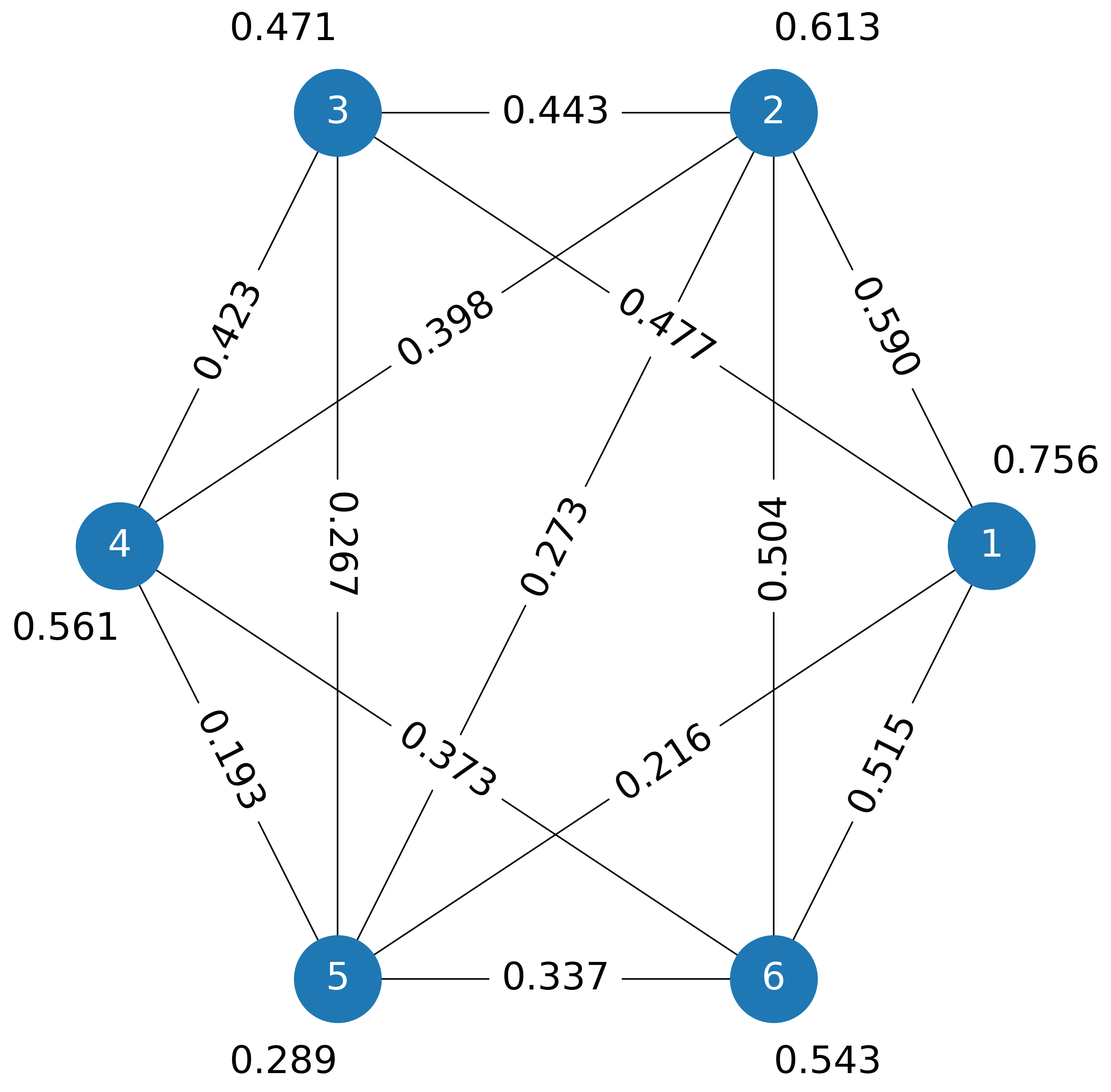}
    \caption{Inter- and intra-annotator agreement scores, whith edges representing pairwise inter-annotator agreement between two annotators and the values next to nodes representing the intra-annotaor agreement score, using a multi-label variant of Krippendorff's alpha. The reliability scores using $\alpha = 0.5$ in the EffiARA reliability score calculation are (from annotator 1 to 6): 1.309, 1.141, 0.945, 0.978, 0.576, 1.051.}
    \label{fig:iaa}
\end{figure}

Inter-annotator agreement can be assessed as each annotator shares their samples evenly with four other annotators. This provides sufficient overlap to understand the pairwise inter-annotator agreement between; this is visualised in Figure \ref{fig:iaa}. To assess intra-annotator agreement, we assign 20 duplicate samples to each annotator. Here, we diverge slightly from \citet{cook2024efficient} by randomly sampling from the complete set of an annotator's annotations, rather than from the set of single-annotated samples (our dataset does not contains only double-annotated samples). Sampling in this way highlights the potential need for the EffiARA sample-distribution algorithm \citep{cook2024efficient} to be generalised to situations where single annotations do not exist or where the user may not wish to limit their sample size for re-annotations.

In the annotation process, we used the EffiARA annotator reliability score to filter out an annotator who had a significantly lower reliability score than all other annotators. The annotations that were removed based on this were then annotated by the most reliable annotator based on the EffiARA reliability score.  

By using this annotation framework and maintaining meta-data about which annotator has annotated each sample, we were able to replace an unreliable annotator and leave scope for any experiments that may utilise such meta-data. \\

\noindent \textbf{Annotator Agreement and Reliability} \\
Each annotator's average inter-annotator agreement, intra-annotator agreement, and reliability factor is calculated as described in \citet{cook2024efficient}. 

Figure \ref{fig:iaa} shows the annotator agreement within the dataset. Each node represents an annotator, the edges represent the pairwise inter-annotator agreement, and the intra-annotator agreement value is displayed next to each node. The agreement metric used in this study was the mean of the Krippendorff's alpha metric for each of the 14 news frames; for each pairwise agreement calculation, we calculate the agreement between two users per class and divide by the number of classes (or news frames). The average Krippendorff's alpha of all links (excluding annotations from annotator 5, the least reliable annotator, as their annotations did not contribute towards the gold-standard) is 0.465. While this is lower than the recommended level of 0.667, it is higher than that of \citet{piskorski-etal-2023-semeval}, who report an inter-annotator agreement of 0.342 in their dataset. 

Note that there is an overlapping set of annotations between annotator 2 and annotator 5, shown in Figure \ref{fig:iaa}. This is due to the reliable annotator 2 taking the place of the less reliable annotator 5. \\

\noindent \textbf{Annotation Adjudication} \\
Once all samples had been double-annotated, the senior annotator reviewed the annotations, resolving any identified conflicts as suggested in \citet{piskorski-etal-2023-semeval}. Any overlapping labels are automatically considered golden standard, with the senior annotator using their discretion for other classes. This adjudication process produced a set of annotations considered to be the golden standard.

\subsection{Chinese News Framing Dataset}
The Chinese News Framing dataset consists of 353 Chinese articles annotated into at least one news frame. In general, the dataset consists of 172 and 181 news articles from 14 news sources respectively. Following the structure of the SemEval dataset \citep{piskorski-etal-2023-semeval}, the Chinese Framing dataset is split into three subsets: training set (233), development set (50) and test set (70). 

In Table \ref{tab:dataset_splits}, we present detailed descriptive statistics for the three subsets and the full dataset, including the number of samples per domain, year, frame, etc. For reference, we also provide descriptive statistics of our dataset with the SemEval news framing dataset \citep{piskorski-etal-2023-semeval} in Table \ref{tab:statistics}.

\begin{table*}[!t]
\resizebox{\textwidth}{!}{%
\begin{tabular}{lllllllllll}
\hline
\rowcolor[HTML]{EFEFEF} 
\textbf{Metric} &
  \textbf{Chinese} &
  \textbf{English} &
  \textbf{French} &
  \textbf{German} &
  \textbf{Italian} &
  \textbf{Polish} &
  \textbf{Russian} &
  \textbf{Georgian} &
  \textbf{Greek} &
  \textbf{Spanish} \\ \hline
\rowcolor[HTML]{EFEFEF} 
\textbf{\#docs} & \textbf{353} & 590 & 261 & 227 & 364 & 241 & 263 & 29 & 64 & 30 \\ 
\rowcolor[HTML]{EFEFEF}
\textbf{\#avg. Frames} & \textbf{3.1} & 4.0 & 3.0 & 4.7 & 3.8 & 5.2 & 2.1 & 1.7 & 2.9 & 2.3 \\ 
\textbf{1: Economic} & \textbf{111} & 74 & 79 & 108 & 142 & 144 & 68 & 2 & 14 & 4 \\ 
\textbf{2: Capacity \& Resources} & \textbf{102} & 56 & 62 & 104 & 120 & 88 & 34 & 4 & 10 & 44 \\ 
\textbf{3: Morality} & \textbf{35} & 231 & 62 & 39 & 62 & 63 & 31 & 2 & 5 & 7 \\ 
\textbf{4: Fairness \& Equality} & \textbf{42} & 131 & 30 & 35 & 52 & 39 & 21 & 0 & 8 & 2 \\ 
\textbf{5: Legality \&
Jurisprudence} & \textbf{75} & 281 & 41 & 65 & 73 & 56 & 44 & 0 & 23 & 7 \\ 
\textbf{6: Policy Prescription} & \textbf{168} & 154 & 38 & 70 & 129 & 110 & 15 & 2 & 12 & 7 \\ 
\textbf{7: Crime \& Punishment} & \textbf{86} & 274 & 22 & 44 & 57 & 57 & 51 & 3 & 11 & 4 \\ 
\textbf{8: Security \& Defense} & \textbf{113} & 222 & 89 & 121 & 155 & 105 & 90 & 10 & 19 & 10 \\ 
\textbf{9: Health \& Safety} & \textbf{62} & 86 & 60 & 107 & 97 & 144 & 37 & 4 & 8 & 3 \\ 
\textbf{10: Quality of Life} & \textbf{61} & 115 & 40 & 53 & 89 & 85 & 32 & 0 & 5 & 3 \\ 
\textbf{11: Cultural Identity} & \textbf{39} & 42 & 34 & 46 & 43 & 48 & 13 & 1 & 8 & 0 \\ 
\textbf{12: Public Opinion} & \textbf{48} & 68 & 34 & 50 & 58 & 74 & 22 & 4 & 10 & 3 \\ 
\textbf{13: Political} & \textbf{45} & 343 & 108 & 130 & 178 & 144 & 55 & 10 & 43 & 6 \\ 
\textbf{14: External Reputation} & \textbf{120} & 214 & 85 & 91 & 132 & 86 & 44 & 9 & 9 & 3 \\ \hline
\end{tabular}%
}
\caption{Descriptive statistics between the  Chinese Framing dataset (see bold values in the first column) and languages from the SemEval dataset \citep{piskorski-etal-2023-semeval}.}
\label{tab:statistics}
\end{table*}

\section{Experimental Setup}
In this section, we present experiments designed to evaluate the quality of our dataset and the value of its integration into the existing SemEval dataset. 

The aims of our experiments are as follows:
\begin{enumerate}[label=(\roman*)]
    \item provide baseline results for each language when models are fine-tuned in a multilingual setting, allowing for the assessment of our Chinese News Framing dataset's value;
    \item provide baseline results for each language using a state-of-the-art decoder model (GPT-4o \citep{openai2024gpt4technicalreport}) for comparison with fine-tuned XLM-RoBERTa models;
    \item understand whether our Chinese news framing dataset facilitates the creation of monolingually fine-tuned models capable of classifying articles into a set of given news frames;
    \item understand the impact of augmenting the SemEval dataset with Chinese News Framing samples using the classification performance of each language as the performance metric.
\end{enumerate}

\subsection{Augmenting the SemEval Dataset}
As shown above, our dataset was developed following a similar methodology of that used to create the SemEval dataset \citep{piskorski-etal-2023-semeval}. This suggests that our dataset can not only be used as a stand-alone resource for Chinese news analysis but also as a complementary addition to the SemEval dataset.

To assess the value of our dataset, we experiment with three training sets, the original SemEval dataset, our novel Chinese News Framing dataset, and the augmented SemEval dataset containing Chinese news framing samples. In these experiments, we treat the SemEval development set as the test set due to the official SemEval test set not being publicly available. We consider all the available languages offered in the SemEval train and dev sets: English, French, German, Italian, Polish, and Russian. Statistics for the augmented SemEval dataset are shown in Table \ref{tab:statistics}.

\subsection{Model}

We conduct experiments using XLM-RoBERTa-base\footnote{\url{https://huggingface.co/FacebookAI/xlm-roberta-base}} \citep{conneau2020unsupervised}. XLM-RoBERTa was one of the most widely used transformer-based multilingual models among SemEval participants \citep{piskorski-etal-2023-semeval, jiang-2023-team, liao-etal-2023-marseclipse}.

\subsection{Experimental Details}
\noindent \textbf{Text Pre-processing} \\
To prepare the article text for classification, following \citet{wu2023sheffieldveraai} we conduct the following additional pre-processing steps:

\begin{itemize} \item a newline character is added between the news title and body;
\item duplicate sentences occurring consecutively are removed;
\item hyperlinks to websites and images are removed;
\item strings detailing author biographies (such as names and affiliations) are also removed.
\end{itemize}

\noindent \textbf{Model Training} \\
We fine-tune XLM-RoBERTa-base for 100 epochs, with 10 warm-up epochs, a batch size of 8, and a max-sequence length of 512 tokens. We utilise the AdamW optimiser with a linear weight decay of 0.01. The loss function used in this multi-class, multi-label classification task was Binary Cross-Entropy with Logit Loss.

The learning rate hyperparameter was selected through initial experimentation on the Chinese news framing development set, as our dataset offers train, development, and test splits. These initial experiments involved fine-tuning over only 30 epochs, maintaining a warmup rate of 0.1. The learning rates tested were those used by \citet{wu2023sheffieldveraai} across their experiments: 1e-4, 5e-5, 3e-5, 2e-5, and 5e-6. The best-performing learning rate, measured by the F1-micro score on the Chinese News Framing development set, was selected for the final set of experiments. 

All models are trained three times with the set of three seeds \{555, 666, 777\}, using the highest performing learning rate obtained using the Chinese-only development set as described above. The key performance metric in this study, following the official performance metric of SemEval-2023 task 3 subtask 2 \citep{piskorski-etal-2023-semeval}, is the F1-micro score; we report the mean across the three seeds and the standard deviation for each language. The code and configuration files required to run all experiments all experiments are available at \url{https://github.com/GateNLP/chinese-news-framing}.

All experiments were run on an Nvidia RTX 4090 with 24GB of VRAM. \\

\noindent \textbf{GPT-4o Experiments} \\
We employ GPT-4o \citep{openai2024gpt4technicalreport} in a zero-shot setting to generate frame labels in a comma-separated format. The temperature parameter is set to 0.0 for deterministic outputs. Following our supervised experiments, we conduct three runs per language and report the mean F1-micro score with standard deviation.

\section{Results and Discussion}

\begin{table*}[!t]
\centering 
\begingroup 
\renewcommand{\arraystretch}{1.5} 
\resizebox{\textwidth}{!}{%
\begin{tabular}{cccccccc}
\toprule
\textbf{Training Data} &
  \textbf{Chinese} &
  \textbf{English} &
  \textbf{French} &
  \textbf{German} &
  \textbf{Italian} &
  \textbf{Polish} &
  \textbf{Russian} \\ 
\midrule
\textbf{Zero-Shot ChatGPT 4o}& 
0.560±0.004 & 
0.603±0.009 & 
0.528±0.001 & 
0.541±0.004 & 
0.540±0.002 & 
0.575±0.010 & 
0.508±0.002 \\ 
\textbf{SemEval} \citep{piskorski-etal-2023-semeval} & 
0.584±0.016 & 
0.733±0.011 & 
\textbf{0.589±0.017} & 
\textbf{0.643±0.014} &
\textbf{0.599±0.011} & 
0.647±0.009 & 
\textbf{0.584±0.014} \\ 
\textbf{Chinese News Framing} & 
0.719±0.012 & 
0.570±0.013 & 
0.433±0.011 & 
0.515±0.013 & 
0.546±0.003 & 
0.516±0.016 & 
0.434±0.024 \\ 
\textbf{SemEval + Chinese News Framing} &
\textbf{0.753±0.015} & 
\textbf{0.739±0.023} & 
0.578±0.007 &
0.639±0.023 &
0.592±0.004 &
\textbf{0.670±0.007} &
0.542±0.008 \\ 
\bottomrule
\end{tabular}%
}
\endgroup
\caption{Experimental results for GPT-4o in the zero-shot setting and for XLM-Roberta-base fine-tuned with three different training sets, trained for 100 epochs, with learning rate 5e-5.}
\label{tab:roberta-results}
\end{table*}

Table \ref{tab:roberta-results} displays the experimental results of GPT-4o in the zero-shot setting and the results of XLM-Roberta-base being fine-tuned on the three different datasets: the SemEval training set, our Chinese News Framing training set, and the SemEval training set augmented with Chinese News Framing samples.\\

\subsection{Experimental Results}

\noindent \textbf{SemEval Baseline Performance} \\
Our baseline results, using XLM-Roberta-Base fine-tuned on the SemEval training set and tested over three random seeds on the development set closely align with the results obtained by \citet{wu2023sheffieldveraai}, with the only results differing by more than an F1-micro score of over $0.01$ being English, Polish, and Russian which have differences of $+0.05$, $-0.018$, and $+0.035$, where a positive result indicates our model performs better. It is worth noting the difference in experimental setup. The experiments conducted by \citet{wu2023sheffieldveraai} involve a 3-fold cross-validation on the development set to create three different test sets, whereas we maintain the same test set (the SemEval development set) over three different random seeds. 

While not trained on Chinese news framing samples, the model performs within $0.005$ of the scores achieved for French and Russian. The F1-micro achieved on the Chinese test set without explicit Chinese-language fine-tuning was $0.584$. \\

\noindent \textbf{Chinese News Framing Only} \\
Training on the Chinese News Framing dataset, as expected, significantly increases the model's performance on Chinese news framing test samples. Through training on only the Chinese News Framing data, the Chinese test score increases to $0.719$; this is a higher performance than all languages other than English in our SemEval baseline performance results. This positive performance on our dataset indicates that it is a valuable tool and benchmark for the Chinese news framing task. \\

\noindent \textbf{Augmented SemEval Performance} \\
After augmenting the SemEval dataset with our Chinese News Framing dataset, we see a further increase in performance on the Chinese language, achieving an F1-micro of $0.753$ which is the highest score attained on any language in all of our experiments. While most languages have little change (less than $0.01$ in F1-micro) in performance from our SemEval only baseline expriments, the French news framing performance decreases by $0.011$ and Russian by $0.042$; Chinese classification performance increases by $0.171$ and Polish by $0.023$.

By augmenting the SemEval dataset with our Chinese News Framing dataset, we observe very comparable classification performance in the majority of languages with a significant increase in performance in the Chinese language. This indicates that our Chinese News Framing dataset serves as a valuable complementary addition to the SemEval dataset. \\

\noindent \textbf{GPT-4o Performance} \\
The experimental results show that GPT-4o achieves lower F1 scores than XLM-RoBERTa-Base (trained on the augmented SemEval set containing Chinese news framing samples) in all languages. This performance gap arises from the distinct prediction pattern of GPT-4o; despite achieving higher recall, it suffers from significantly lower precision. As a generative large language model, GPT-4o tends to assign more labels to texts, interpreting text-topic associations from an overly broad semantic perspective. This leads to the labelling of content that is only superficially related to target categories. Although this over-generalised approach improves topic discovery, it introduces numerous false positives, ultimately compromising classification accuracy. \\

\subsection{Error Analysis}
We also provide further error analysis on the performance of XLM-RoBERTa-Base trained only on our Chinese News Framing dataset. We analyse the performance of this model to highlight trends on our dataset alone, including potential strengths and weaknesses for the model based on our training set.

\begin{table}[!t]
\small
\centering
\begin{tabular}{lcccc}
\hline
\textbf{Frames} & \multicolumn{1}{c}{\textbf{P}} & \multicolumn{1}{c}{\textbf{R}} & \multicolumn{1}{c}{\textbf{F1}} & \multicolumn{1}{c}{\textbf{Support}} \\ \hline
1: Economic  & 0.83 & 0.90 & 0.86 & 21 \\
2: Capacity \& Resources  & 0.47 & 0.47 & 0.47 & 17 \\
3: Morality  & 1.00 & 0.71 & 0.83 & 7 \\
4: Fairness \& Equality  & 0.71 & 0.71 & 0.83 & 7 \\
5: Legality  & 0.26 & 0.62 & 0.37 & 8 \\
6: Policy  & 0.69 & 0.71 & 0.70 & 28 \\
7: Crime \& Punishment  & 0.94 & 0.73 & 0.82 & 22 \\
8: Security \& Defence  & 0.82 & 0.78 & 0.80 & 23 \\
9: Health \& Safety  & 1.00 & 0.67 & 0.80 & 15 \\
10: Quality of Life  & 0.83 & 0.50 & 0.62 & 10 \\
11: Cultural Identity & 0.88 & 0.70 & 0.78 & 10 \\
12: Public Opinion & 0.38 & 0.27 & 0.32 & 11 \\
13: Political & 1.00 & 0.85 & 0.92 & 13 \\
14: External Regulation & 0.84 & 0.84 & 0.84 & 25 \\ \hline
\end{tabular}
\caption{Classification report on the Chinese News Framing dataset, using XLM-RoBERTa-Base trained exclusively on our dataset. Shows the precision, recall, F1 metric and the support (or number of samples) for each class.}
\label{tab:classifications_report}
\end{table}

\paragraph{Classification Report} Table \ref{tab:classifications_report} shows the precision, recall, F1, and support for each individual class within the Chinese News Framing test set. We observe that the model has particular issues in both precision and recall for the frames (2) Capacity \& Resources, (5) Legality, and (12) Public Opinion. When comparing this to the same metrics for the English language development samples in the SemEval dataset, we also observe that these frames are difficult for the model to classify. A key exception to this is in the English and French language, where (5) Legality is well-classified with respect to the other frames for that language. For English, this can be explained by the large set of samples containing the frame Legality \& Jurisprudence. The frames that are most correctly identified are (1) Economic, (3) Morality, (4) Fairness \& Equality, (13) Political, and (14) External Regulation. This does not align with any of the other languages that are trained only on the SemEval set and does not correspond to the number of samples supporting each class.

It is worth noting that with the Augmented SemEval set, containing Chinese News Framing samples, similar trends are observed across languages. For the Chinese language though, the performance of the model in identifying the frame (12) Public Opinion increases by an F1 score of 0.35. Through the analysis of each languages F1-macro performance, the Chinese language performs significantly better than other languages achieving 0.74; the next best performing language, based on F1-macro, is Polish with a score of 0.54. This indicates that our dataset is particularly good for training models to successfully identify the majority of news frames as well as those with higher support (indicated by the high F1-micro).

\paragraph{Co-occurrence Error Matrix}
\begin{figure}[!t]
    \centering
    \includegraphics[width=0.99\columnwidth]{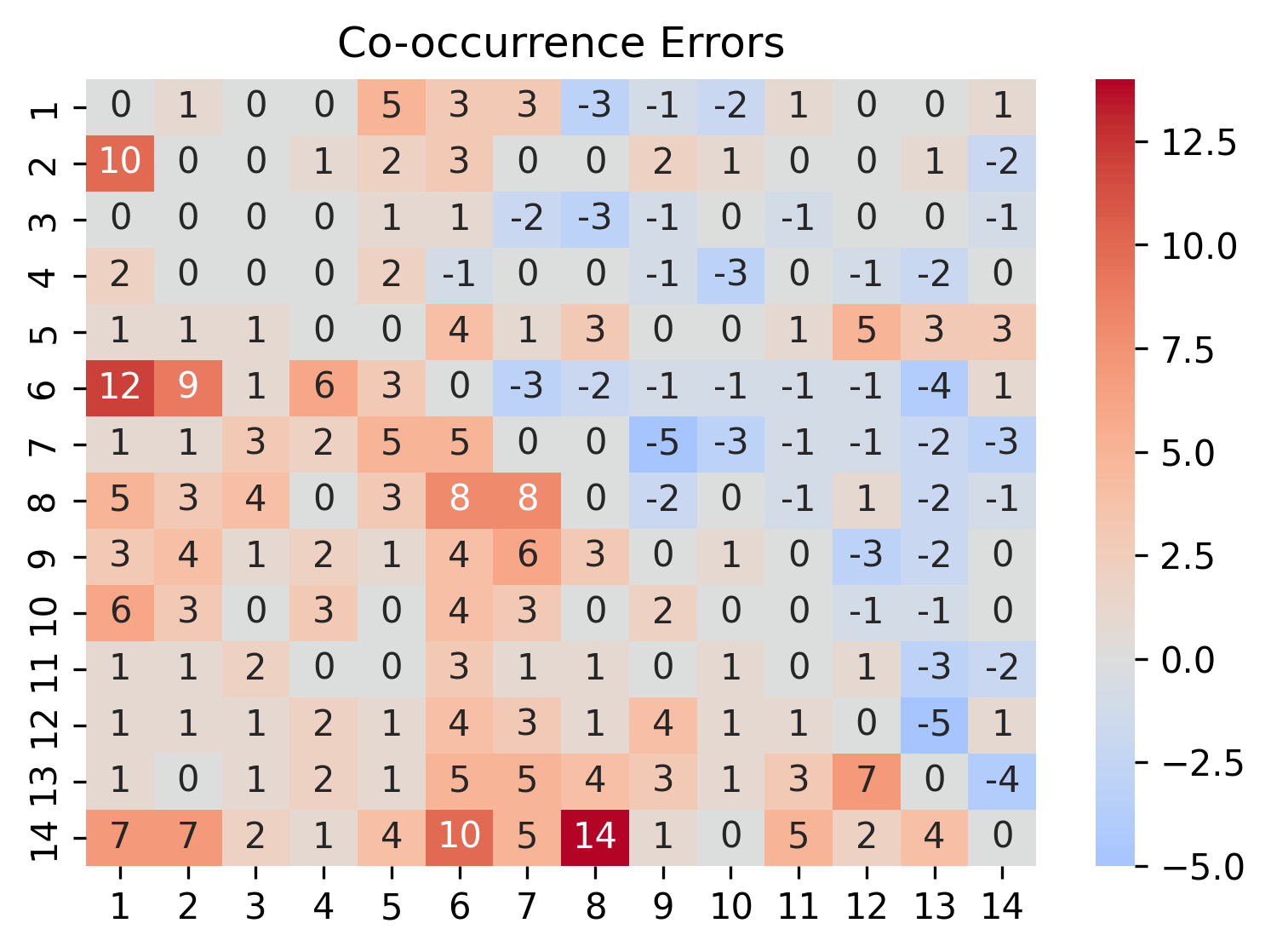}
    \caption{Co-occurrence Error Matrix. Note that labels 1 to 14 denote the news frame orders as in Section `Task Description'. Positive values represent over-prediction and negative values represent under-prediction of the co-occurrence of two classes.}
    \label{fig:co-orr}
\end{figure}
We also conducted co-occurrence error analysis, shown in Figure \ref{fig:co-orr}. The co-occurrence error matrix is used to understand how well the model is able to capture the relationships between pairs of classes in a multi-class, multi-label prediction task. As the co-occurrence matrix is symmetric, we display the number of true co-occurrences in the lower triangle and the difference in co-occurrences in the upper triangle; positive scores indicate that the model has over-predicted the co-occurrence of two news frames and negative scores indicate an under-prediction.

One class showing a number of over-predictions with other classes is (5) Legality. It is particularly over-estimated to co-occur with (1) Economic, (6) Policy, (12) Public Opinion, and (13) Political. This is likely due to the poor precision identified for the (5) Legality news frame shown in Table \ref{tab:classifications_report}.

The (13) Political news frame is shown to be under-predicted to co-occur with a number of news frames, including: (6) Policy, (11) Cultural Identity, (12) Public Opinion, and (14) External Reputation. As the (13) Political news frame is successfully identified by the model but not very well predicted with the correct co-occurring news frames, it indicates that the model does not capture the dependency between news frames. Politics often influences policy, cultural identity, public opinion, and external reputation, but this is not captured very well by the model. Capturing this relationship between news frames in the classification stage may be be of interest in future research.

\section{Conclusion}
In this work, we have created and published the Chinese News Framing dataset, facilitating news framing experiments in the Chinese language and serving as a complementary asset to the SemEval dataset \cite{piskorski-etal-2023-semeval}. We have demonstrated the need for Chinese news framing samples to be added to the SemEval set in order to effectively detect news frames in the Chinese language. We have also reported benchmark results, training on only our Chinese News Framing dataset and an augmented SemEval dataset containing Chinese news framing samples, achieving an F1-micro score of 0.753 on our Chinese News Framing test set; this score is also an improvement over training on only Chinese News Framing samples. Augmenting the SemEval dataset maintains similar F1-micro performance on the SemEval development set, improving the news framing detection in the Chinese, English, and Polish languages.

While this work provides experiments to prove this dataset's efficacy as an addition to the SemEval set, it also provides experiments, achieving high classification performance as a stand-alone dataset. With the extra meta-data available about individual annotations, this work allows for future research involving the development of advanced classification methods, utilising metrics such as annotator reliability \cite{cook2024efficient}.

\section{Limitations}
The articles within this dataset have been collected between 2020 and 2024, offering news framing samples concerning a number of topics. While this serves as a valuable resource, it does not allow for the analysis of news framing over time, allowing for future work to supplement this dataset with the annotation of articles from different periods of time.

As the text is not directly made publicly available in the dataset, relying on the user retrieving content from a web link, there is the risk that some article texts may become unavailable after the release of the dataset.

Although we provided annotator guidelines and training, subjective interpretations of news frames are likely to be present in the dataset. We attempt to mitigate this by providing each individual annotation in our dataset, rather than only the aggregated ``gold-standard'' labels. This allows for further research investigating the uncertainty and subjectivity involved in this task.

\section{Dataset Availability}
To maximise the use of our dataset, we adhere to the FAIR principles \citep{wilkinson2016fair}.
\begin{itemize} 
\item \textit{Findable.} Our dataset has been published to the Zenodo dataset sharing service: 10.5281/zenodo.14659362; our experimental code is also available at https://github.com/GateNLP/chinese-news-framing.
\item \textit{Accessible.} All data is publicly accessible through web links in the published dataset; we also provide a tool to obtain the news article text from the given link in our code repository.
\item \textit{Interoperable.} The CSV format is widely accepted and can be processed by a wide range of data processing tools.
\item \textit{Reusable.} The CSV format is widely accepted and can be processed by a wide range of data processing tools.
\end{itemize}

\section{Ethics Statement}
Our study has received ethical approval from The University of Sheffield. All participants provided informed consent by signing the consent form after reviewing the information sheet detailing the annotation task. Annotator identities remain anonymous and any news content itself will not be contained within the published dataset.

The potential use of this dataset by malicious actors has been considered, potentially allowing the generation of maliciously framed content; this behaviour is already possible with existing Large Language Model technology \citep{gururangan2020don}. Proper use, however, has much more potential for good in the form of understanding media bias, providing balanced search results, and large-scale content analysis in research.

\bibliography{aaai25}

\appendix

\end{document}